# Discussion among Different Methods of Updating Model Filter in Object Tracking

Taihang Dong*[a,b], Sheng Zhong[a,b]
[a]*Science & Technology on Multispectral Information Processing Laboratory;*
[b]*School of Automation, Huazhong University of Science & Technology, Wuhan 430074, Hubei*

## ABSTRACT

Discriminative correlation filters (DCF) have recently shown excellent performance in visual object tracking area. In this paper, we summarize the methods of updating model filter from discriminative correlation filter (DCF) based tracking algorithms and analyzes similarities and differences among these methods. We deduce the relationship between updating coefficient in high dimension (kernel trick), updating filter in frequency domain and updating filter in spatial domain, and analyze the difference among these different ways. We also analyze the difference between the updating filter directly and updating filter's numerator (object response power) with updating filter's denominator (filter's power). The experiments about comparing different updating methods and visualizing the template filters are used to prove our derivation.

## 1. INTRODUCTION

Visual tracking is one of the most basic problems in computer vision with various applications in video surveillance, human-computer interaction and vehicle navigation. Its goal is to localize the object position in continuous image sequences. Although this area has made great progress, it has great potential to get developed facing some particular problems such as geometric deformations, partial occlusions and illumination changes.

Recently, discriminative correlation filter (DCF) based visual trackers [1] [2] [3] have shown to provide excellent performance. They use image patches information in last frames to train the filter and find the maximum of response from the correlation between the filter and image patch in the current frame.

There are various ways to update the model filter. Bolme et al. [4] propose a kind of algorithm named ASEF (Average of Synthetic Exact Filters) filter, whose filter is updated directly with weight:

$$H_i^* = \eta \frac{G_i \odot F_i^*}{F_i \odot F_i^*} + (1-\eta) H_{i-1}^* . \tag{1}$$

Here, $\frac{G_i \odot F_i^*}{F_i \odot F_i^*}$ is filter trained only using image patch in current frame, while $H_i^*$ denotes ASEF filter in frame $i$ . $\eta$ represents learning factor or model updating rate.

Then, Bolme et al. [1] propose MOSSE (Minimum Output Sum of Squared Error) filter, which is predecessor of almost all DCF based algorithms. The updating process renews the filter by formula:

$$H_i^* = \frac{A_i}{B_i} \tag{2}$$

$$A_i = \eta G_i \odot F_i^* + (1-\eta) A_{i-1} \tag{3}$$

$$B_i = \eta F_i \odot F_i^* + (1-\eta) \mathrm{B}_{i-1} . \tag{4}$$

Here, $A_i$ is the numerator of the filter, and $B_i$ is denominator of the filter. The MOSSE algorithm updates the numerator and the denominator of the filter respectively in the same learning factor.

Conceptually, the first successful theoretical extension of the standard DCF was the kernelized formulation by Henriques et al. [2] [3]. They analyze the formula generation in theory and give the interpretation that DCF based tracking algorithm essentially provides many virtual samples to train the filter. Meanwhile, they generalize the algorithm to high dimension by kernel trick. The filter updating is performed directly:

$$\hat{M}^{p} = (1-\eta)\hat{M}^{p-1} + \eta\hat{M}^{curr} \tag{5}$$

$$\hat{\alpha}^{p} = (1-\eta)\hat{\alpha}^{p-1} + \eta\frac{\hat{y}}{\hat{k}^{xx}+\lambda} . \tag{6}$$

*taih.dong@gmail.com

Here, $M$ represents feature extracted from image patch. $\alpha$ represents the solution in the dual space and superscript $p$ and $curr$ are denoted to p-th and current frame. The caret $\hat{\bullet}$ is expressed as in frequency domain and $\frac{\hat{y}}{\hat{k}^{xx}+\lambda}$ is the filter trained in current frame.

DCF algorithms are similar to template-matching algorithms in spatial domain in some degree. The conventional and original correlation method to solve the tracking problem is NCC (normalized cross correlation). NCC updates the image template directly. The template in NCC could be updated by interpolating:

$$T^{p} = (1-\eta)T^{p-1} + \eta T^{curr} . \tag{7}$$

Where $T$ indicates the template.

## 2. THE ANALYSES ABOUT FILTER UPDATING METHODS

There are several patterns of updating filter used in tracking area frequently. They are updating filter in frequency domain, updating filter in spatial domain, updating filter coefficients in high dimension (kernel trick) directly and updating filter fractionally. The difference between the updating filter in spatial domain and frequency domain is first analyzed. Then we deduce the relationship between updating filter in normal dimension and updating filter coefficients in high dimension (kernel trick) directly. Finally, we analyze the difference between updating filter directly and updating filter' numerator (object response power) with updating filter' denominator (filter's power).

### 2.1. The analyses of filter updating methods in different domains

Fourier transform has the linear behavior:

$$ax_1(n) + bx_2(n) \leftrightarrow aX_1(\omega) + bX_2(\omega) . \tag{8}$$

Here, $a$ and $b$ represent the coefficients or weight of the different element in spatial or frequency domain. The lower case letter $x(n)$ is signal in spatial domain while capital letter $X(\omega)$ is expressed as in frequency domain.

So the sum of the frequency spectrum with weight corresponds to the sum of the spatial image patches with the same weight. Feature summation and subtraction in spatial domain is equal to addition and subtraction of the respective element in frequency domain. That means updating filter in spatial domain and in frequency domain have the same effect.

$$x^{p} = (1-\eta)x^{p-1} + \eta x^{curr} \Leftrightarrow X^{p} = (1-\eta)X^{p-1} + \eta X^{curr} . \tag{9}$$

Here, $X$ is frequent form of feature or signal $x$.

### 2.2. The analyses of filter updating methods with and without kernel trick

With kernel functions in tracking to separate the object positive samples and negative samples, ways of updating filter could be various. The most common methods are to update the coefficient of high dimension feature or signal and update model template abstracted from the object patch directly.

For universal property and convenience, The Gaussian function is used as kernel function to analyze the difference among diverse filter updating strategies. The Gaussian kernel is:

$$\kappa(x, x') = \exp(-\frac{1}{\sigma^2}\|x - x'\|^2) . \tag{10}$$

Here, $x$ and $x'$ are different feature vectors. We could tell Gaussian function is obviously a kind of radial basis function. Kernel correlation of two arbitrary vectors, $x$ and $x'$, is the vector $k^{xx'}$ with elements:

$$k_i^{xx'} = \kappa(x', P^{i-1}x) . \tag{11}$$

Here, $P$ is cyclic shift operator. We unfold the Gaussian function and get the expression of vector $k^{xx'}$:

$$k^{xx'} = \exp(-\frac{1}{\sigma^2}(\|x\|^2 + \|x'\|^2 - 2\mathcal{F}^{-1}(\hat{x}^* \odot \hat{x}'))) . \tag{12}$$

In this formula, $\mathcal{F}^{-1}$ represents inverse Fourier transformation and $\odot$ stands for dot product. Then, we substitute $\hat{k}^{xx'}$ in $\frac{\hat{y}}{\hat{k}^{xx} + \lambda}$ to compute the coefficient $\hat{\alpha}$ in high dimension:

$$\hat{\alpha} = \frac{\hat{y}}{\mathcal{F}(\exp(-\frac{1}{\sigma^2}(\|x\|^2 + \|x\|^2 - 2\mathcal{F}^{-1}(\hat{x}^* \odot \hat{x})))) + \lambda} . \tag{13}$$

Here, $\hat{y}$ is desired output in frequency domain and $\mathcal{F}$ represents Fourier transformation. $\lambda$ is the constant on behalf of regularity term coefficient. Both methods, updating coefficient $\hat{\alpha}$ directly and updating feature in frequency domain, can update the parameter to adapt the changing of object appearance. The updating rates of these two methods are different. We expect to compute the partial derivatives $\frac{\partial \hat{\alpha}}{\partial \hat{x}}$ of Eq. (13) with respect to $\hat{x}$, but the fact is that there is no closed-form solution. So we only consider Gaussian function part $\mathcal{F}(\exp(-\frac{1}{\sigma^2}(\|x\|^2 + \|x\|^2 - 2\mathcal{F}^{-1}(\hat{x}^* \odot \hat{x}))))$ to find the relationship between Gaussian function and $\hat{x}$. Updating coefficient $\hat{\alpha}$ directly essentially updates the reciprocal of $\hat{k}^{xx}$. We get:

$$\begin{aligned}\hat{k}^{xx} &= \mathcal{F}(\exp(-\frac{1}{\sigma^2}(\|x\|^2 + \|x\|^2 - 2\mathcal{F}^{-1}(\hat{x}^* \odot \hat{x})))) \\ &= \mathcal{F}(\exp(-\frac{2}{\sigma^2}\|x\|^2) \times \exp(\frac{2}{\sigma^2}\mathcal{F}^{-1}(\hat{x}^* \odot \hat{x}))) .\end{aligned} \tag{14}$$

If we assume $\Delta\hat{x} \to 0$, the expression could be simplified:

$$\hat{k}^{xx} = \exp(-\frac{2}{\sigma^2}\|x\|^2) \times \mathcal{F}(\exp(\frac{2}{\sigma^2}\mathcal{F}^{-1}(\hat{x}^* \odot \hat{x}))) . \tag{15}$$

The operator of $\hat{k}^{xx}$ maps the convolution of features $\mathcal{F}^{-1}(\hat{x}^* \odot \hat{x})$ through the exponential function, and zooms the frequency-domain expression of convolution.

That indicates updating coefficient $\hat{\alpha}$ directly is equal to executing a series of mapping, scaling and reciprocal operator basing on updating $\hat{x}$. No matter which kind of method is chosen to update filter, they all converge to coefficients or feature representing the current object state.

### 2.3. The analyses of fractional filter updating and direct filter updating

Filter needs to quickly adapt in order to follow the object while tracking. Running average is used for robustness usually. We use the updating strategies from ASEF and MOSSE as the example to analyze the difference between the fractional filter updating and direct filter updating.

For clear compare, we simplify the formulas from (1) (2) (3) (4) here. The updating strategy in ASEF is shown as:

$$\frac{A_i}{B_i} = \eta\frac{A_{new}}{B_{new}} + (1-\eta)\frac{A_{i-1}}{B_{i-1}} . \tag{16}$$

Here, $A_{new} = G_i \odot F_i^*$ and $B_{new} = F_i \odot F_i^*$. $\frac{A}{B}$ represents filter and $\eta$ is learning rate. The subscript $i$ and $new$ are expressed as frame number and current frame respectively. Then, the updating equation in MOSSE is:

$$\frac{A_i}{B_i} = \frac{\eta A_{new} + (1-\eta)A_{i-1}}{\eta B_{new} + (1-\eta)B_{i-1}} . \tag{17}$$

The robustness indicator $R$ is defined to express the robustness of updating strategy:

$$R = \frac{1}{\left|\frac{A_i}{B_i} - \frac{A_{i-1}}{B_{i-1}}\right|}. \tag{18}$$

The robustness indicator $R$ is equal to the reciprocal of filter changing rate. In other words, when the filter changes more dramatically, the tracker is less robust. For both updating strategies, we compute the robustness indicators:

$$R_{ASEF} = \frac{B_{new}B_{i-1}}{\left|\eta(A_{new}B_{i-1} - A_{i-1}B_{new})\right|} \tag{19}$$

$$R_{MOOSE} = \frac{(\eta B_{new} + (1-\eta)B_{i-1})B_{i-1}}{\left|\eta(A_{new}B_{i-1} - A_{i-1}B_{new})\right|}. \tag{20}$$

Here, $R_{ASEF}$ is denoted as robustness indicator of filter updating strategy in ASEF and $R_{MOOSE}$ represents robustness indicator of MOSSE.

The greater the robustness indicator $R$ is, the more robust the updating strategy is. So if $B_{new} > B_{i-1}$, the filter updating strategy of ASEF is more robust than the MOSSE, and vice versa. This analysis declares that the filter updating rate is changing as the energy which the training image contains changes. The robustness of these two strategies is changing as $B$ changes.

## 3. RESULTS AND DISCUSSION

We experiment the effect of different filter updating strategies to prove our derivation in theory. In Section 3.1, we experiment the difference and comparison between the filter updating strategies in spatial and frequency domain. The relationship among various strategies of updating filter with and without kernel function is shown in Section 3.2. Section 3.3 illustrates the comparison between fractional filter updating and direct filter updating.

### 3.1. Comparison between filter updating methods in spatial domain and frequency domain

Our deviation and analysis are tested based on KCF (kernel correlation filter) [2] [3] in OTB dataset [5] [6]. For better visualization, we use HOGgles (HOG goggles) [7] [8] to visualize the filter of HOG feature in spatial domain. As for the feature in frequency domain, we transform it to spatial domain and then visualize it through HOGgles [7] [8].

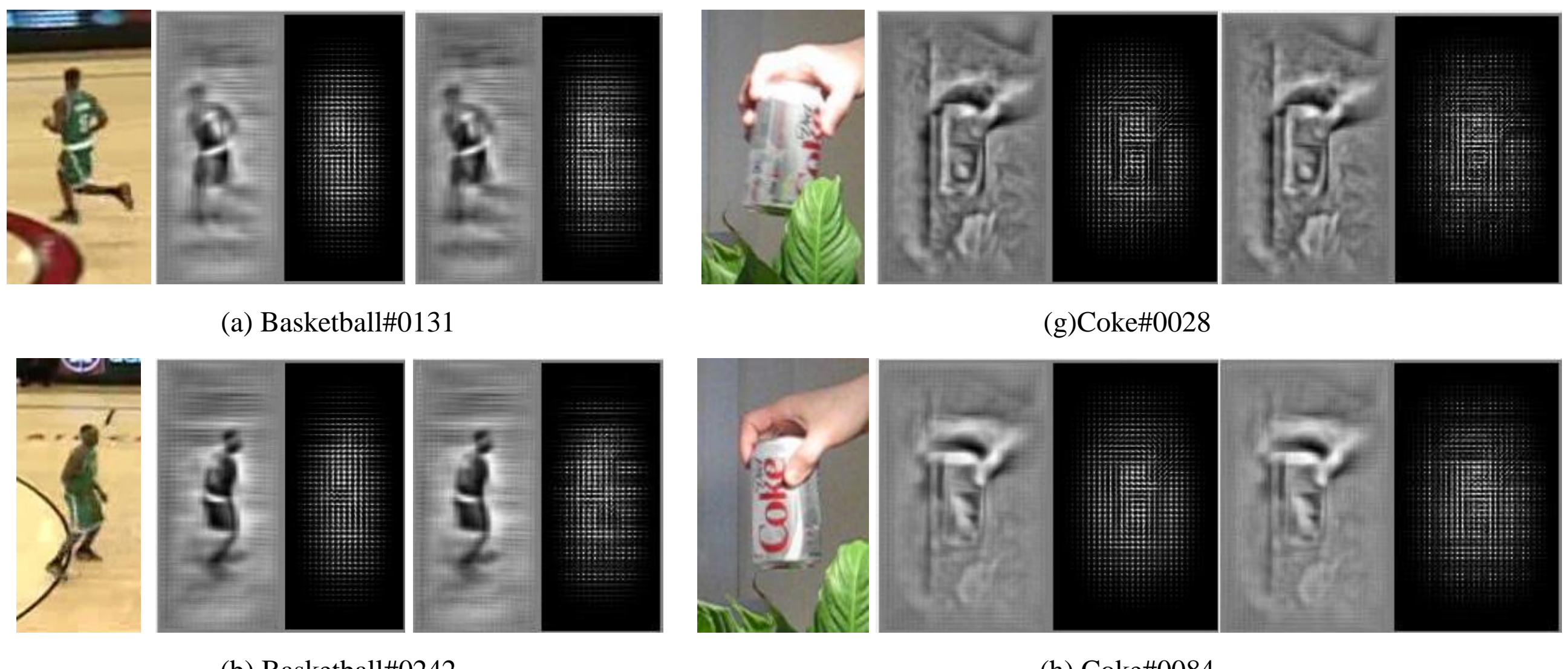

(a) Basketball#0131

(g)Coke#0028

(b) Basketball#0242

(h) Coke#0084

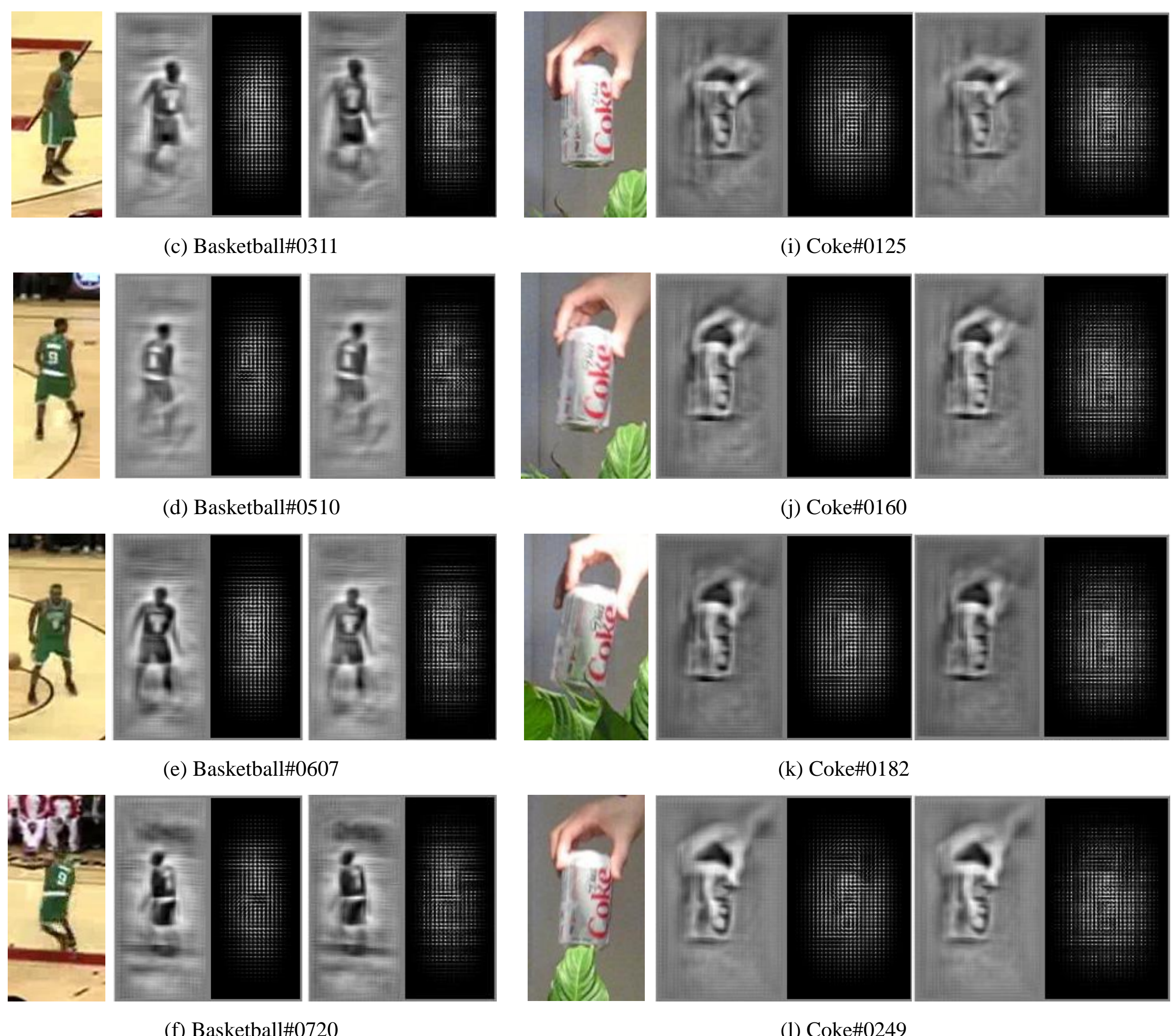

(c) Basketball#0311 (i) Coke#0125

(d) Basketball#0510 (j) Coke#0160

(e) Basketball#0607 (k) Coke#0182

(f) Basketball#0720 (l) Coke#0249

Fig. 1 (a-f) belongs to Basketball video, and (g-l) belongs to Coke video. For each cluster of image patches, the left is original image, the middle picture is template filter updating in spatial domain and the right picture is template filter updating in frequency domain.

We illustrate the typical example videos 'Basketball', 'Coke' and corresponding results in Fig. 1. We could tell updating filter directly has the same effect as updating frequency spectrum of the filter. The similarity of updating in spatial and frequency domain indicates our theory is reasonable. In fact, there is little difference that is hardly noticed between filters using different updating strategies in spatial domain and frequency domain. The reason is that there is some computational error of FFT (fast Fourier transformation) and this kind of error could accumulate to impact performance of the tracker.

### 3.2. Relationship among filter updating methods with and without kernel trick

The variation trend of filter updating strategies with and without kernel trick would be analyzed. For clarity and convenience, we simplify the Eq. (13) to assume that the $x$ is a scalar. We set $x_{init}$ to 2 which is computed from last frames. For different $x_{curr}$ which represents feature in current frame, $x_{upd}$ is computed by interpolating with learning rate $\eta = 0.025$. The standard variance $\sigma$ of Gaussian function is set to 60 and the $\|x\|^2$ is set to the fixed value $x_{init}^2$ for approximation. For convenience, $y$ is set to 1. We simulate the filter changing rate to visualize the relationship between filter updating strategies with and without kernel trick in Fig. 2(a). The abscissa reflects $x_{upd}$ value and ordinate is $\alpha$ value.

The red line in Fig. 2(a) represents the curve of updating feature before kernel operation, which means we update image feature first and then substitute the image feature into kernel function. The green line in Fig. 2(a) is expressed as updating coefficient after the kernel function. Both the lines have the Gaussian distribution essentially, but the red line is smoother. That declares that updating image feature first and then substituting the image feature into kernel function is more robust. The green line is approximate to red line around $x_{init} = 2$. That indicates updating coefficient in high dimension has the similar effect to updating template filter in frequency domain directly while the feature $x_{curr}$ is similar to $x_{init}$, and it doesn't change drastically.

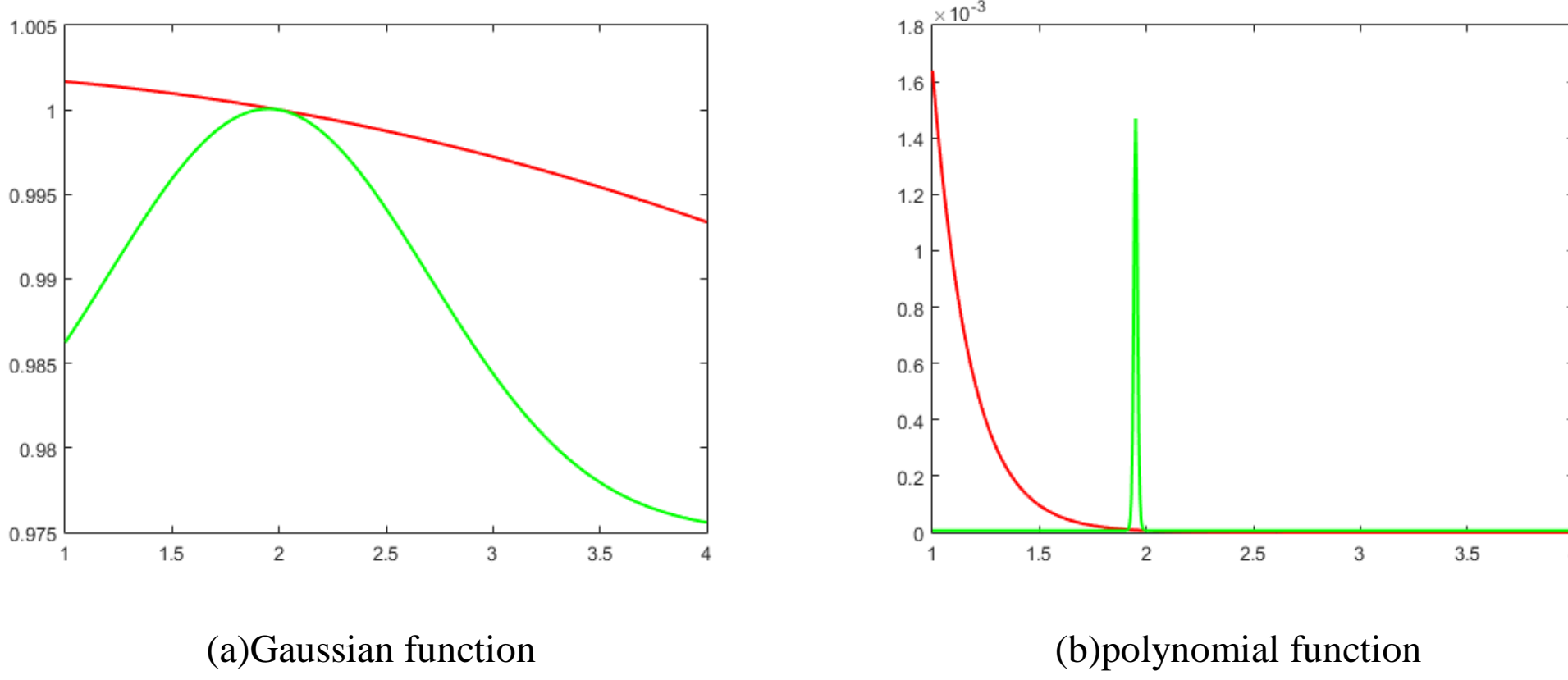

(a)Gaussian function (b)polynomial function

Fig. 2 The simulation about filter updating methods with and without kernel trick. The red line represents filter updating without kernel function and green line with kernel function.

We simulate the filter changing curve using polynomial function as shown in Fig. 2(b). The polynomial kernel additive term is set to 1.5 and the exponent is set to 7. The other setting is same as Gaussian function. We see a sharp peak in Fig. 2(b) less than $x_{upd} = 2$ place. This holds the same phenomenon that updating image feature first and then substituting the image feature into kernel function is more robust. The green line is approximate to the red line around $x_{init} = 2$, so there is similar effect between different updating strategies while feature $x_{curr}$ doesn't change drastically.

Different kernel functions have different effects because of the different curve shapes. The visualization of these different ways of updating proves our analysis.

### 3.3. Comparison about methods of fractional filter updating and direct filter updating

We simplify the comparison in the same way to visualize the filter changing rate in fractional filter updating and direct filter updating. Considering Eq. (16) (17) (18), we set $\frac{A_{i-1}}{B_{i-1}}$ to the fixed value $\frac{1}{2}$, and $A_{new}$ is set to 1. $B_{new}$ changes in interval $\left[1,3\right]$. We illustrate the filter changing rate in Fig. 3(a) and robust indicator in Fig. 3(b). The abscissa in Fig. 3 is $B_{new}$. The ordinate in Fig. 3(a) is filter value and in Fig. 3(b) is robust indicator value.

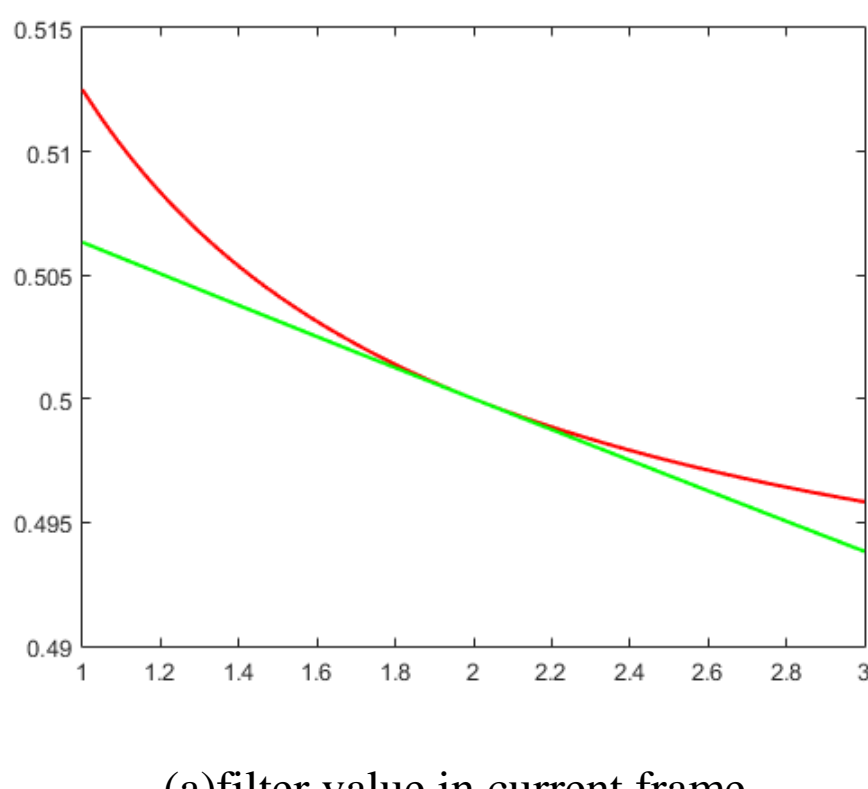

(a)filter value in current frame

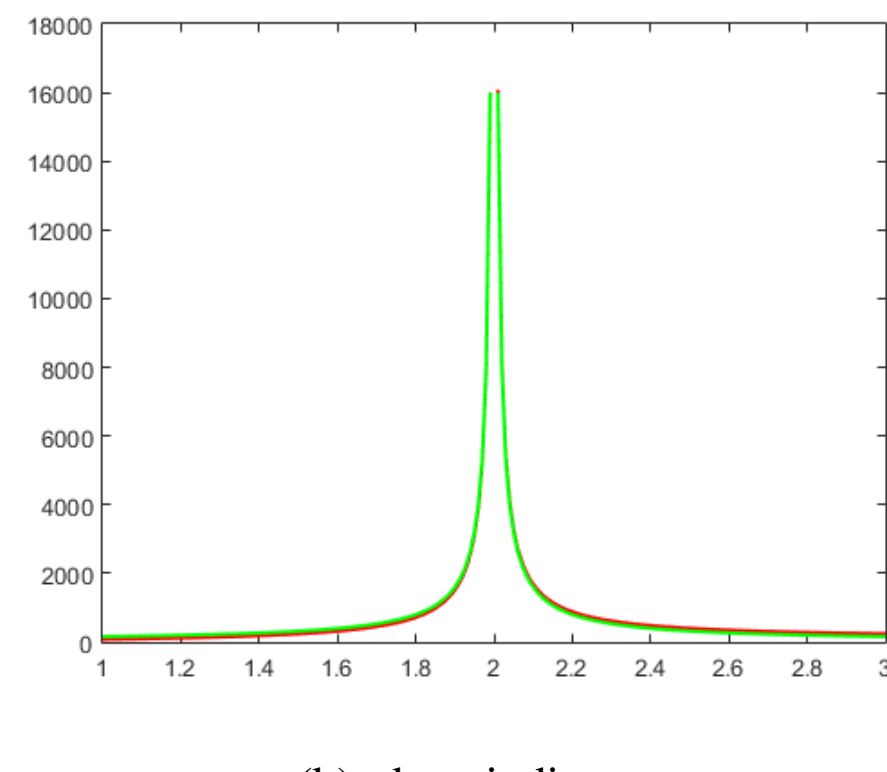

(b)robust indicator

Fig. 3 The simulation about fractional filter updating and direct filter updating. The red line represents direct filter updating and green line fractional filter updating.

We can see when $B_{new} = B_{i-1} = 2$, these two strategies have totally same impact on updating filter. The red line is always above the green line, which means that filter through direct updating is greater than the filter through fractional updating. In Fig. 3(b), the red line is above the green line when $B_{new} > B_{i-1} = 2$, and red line is below the green line when $B_{new} < B_{i-1} = 2$. The legend is consistent with theoretical deduction in Section 2.3.

If the power is fixed that the image patch contains, there is no difference between updating filter directly and updating filter's numerator (object response power) with updating filter's denominator (filter's power). If the power aforementioned changes among different image patches, the filter changing rate is changing. Whether direct updating filter has the faster filter-updating performance depends on whether $B_{new} > B_{i-1} = 2$.

## 4. CONCLUSION

There are various filter updating methods to renovate the template in DCF based tracking algorithms. There are specific relationships between these updating strategies and the experiments prove our derivation. Updating filter in spatial domain is same as updating filter in frequency domain. In general, updating coefficient in high dimension directly changes filter faster. The concrete filter updating rate is relative to the kernel function form. In terms of the difference between updating filter directly and updating object response power with updating filter's power, we define robustness indicator to analyze it and draw the conclusion that whether updating filter directly has the faster filter updating performance depends on whether the power of image patch in the current frame is greater than last frames. If there is little difference between adjacent frames, there is no more difference between these filter updating strategies. Different updating strategies might adapt to different learning rates, so we would explore the best matches between the updating strategies and learning rates in the future.

## ACKNOWLEDGEMENTS

The authors are grateful for the support of the National Key Research and Development Program No.2016YFF0101502.

## REFERENCES

[1]. Bolme, D. S., Beveridge, J. R., Draper, B. A., & Lui, Y. M. (2010). Visual object tracking using adaptive correlation filters. *Computer Vision and Pattern Recognition* (pp. 2544-2550). IEEE Computer Society.

[2]. Henriques, J. F., Caseiro, R., Martins, P., & Batista, J. (2012). Exploiting the circulant structure of tracking-by-

detection with kernels. *European Conference on Computer Vision* (pp. 702-715). Springer, Berlin, Heidelberg.

[3]. Henriques, J. F., Caseiro, R., Martins, P., & Batista, J. (2015). High-speed tracking with kernelized correlation filters. *IEEE Transactions on Pattern Analysis and Machine Intelligence*, *37*(3), 583-596.

[4]. Bolme, D. S., Draper, B. A., & Beveridge, J. R. (2009). Average of synthetic exact filters. *Computer Vision and Pattern Recognition* (pp. 2105-2112). IEEE Computer Society.

[5]. Wu, Y., Lim, J., & Yang, M. H. (2013). Online object tracking: A benchmark. *Computer Vision and Pattern Recognition* (pp. 2411-2418). IEEE Computer Society.

[6]. Wu, Y., Lim, J., & Yang, M. H. (2015). Object tracking benchmark. *IEEE Transactions on Pattern Analysis and Machine Intelligence*, *37*(9), 1834-1848.

[7]. Vondrick, C., Khosla, A., Malisiewicz, T., & Torralba, A. (2013). Hoggles: Visualizing object detection features. *International Conference on Computer Vision* (pp. 1-8). IEEE Computer Society.

[8]. Vondrick, C., Khosla, A., Pirsiavash, H., Malisiewicz, T., & Torralba, A. (2016). Visualizing object detection features. *International Journal of Computer Vision*, *119*(2), 145-158.